\documentclass{ifacconf}

\usepackage{graphicx}      
\usepackage{natbib}        
\begin{document}
\begin{frontmatter}

\title{Pose estimation and bin picking for deformable products} 

\author[First]{Benjamin Joffe} 
\author[First]{Tevon Walker} 
\author[First]{Remi Gourdon}
\author[First]{Konrad Ahlin}

\address[First]{Georgia Tech Research Institute, Atlanta, GA, 30318, USA (e-mail: joffe@gatech.edu, tevon.walker@gtri.gatech.edu, konrad.ahlin@gtri.gatech.edu, remi.gourdon@gatech.edu).}

\begin{abstract}                
Robotic systems in manufacturing applications commonly assume known object geometry and appearance. This simplifies the task for the 3D perception algorithms and allows the manipulation to be more deterministic. However, those approaches are not easily transferable to the agricultural and food domains due to the variability and deformability of natural food. We demonstrate an approach applied to poultry products that allows picking up a whole chicken from an unordered bin using a suction cup gripper, estimating its pose using a Deep Learning approach, and placing it in a canonical orientation where it can be further processed. Our robotic system was experimentally evaluated and is able to generalize to object variations and achieves high accuracy on bin picking and pose estimation tasks in a real-world environment.

\end{abstract}


\end{frontmatter}

\section{Introduction}

Most robot systems that have a perception component and are applied to object manipulation require rather strict assumptions about the object's appearance and geometry. Furthermore, it is common for the locations and poses of the objects to be restricted. Those assumptions can rarely be held in the agricultural and food processing domains: objects have very large variability, complex geometry, and are often deformable. We propose a robotic system that addresses the bin picking problem demonstrated on poultry products. The goal of the system is to take an unordered bin of whole chickens and transform individual objects into a canonical pose, at which point automation can be applied more easily to specific applied tasks. To achieve this, the process is performed in three steps: detect individual objects in a bin and pick them up using a suction cup gripper, perform pose estimation on a singulated object, and place the object in a canonical pose.

Our robotic system consists of a UR-5 robot arm and one Intel RealSense D435 RGB-D camera. To detect the object, we use a standard Faster R-CNN \citep{ren_faster_2015} architecture with Resnet 101 feature exractor. The detected object centroids are converted to 3D and sent to the robot as picking goals. Picking chicken and other natural products is challenging because they deform as they are being picked, they are slippery, and are subject to food safety rules. We utilize a suction cup gripper that aims to make contact with the object at a flat area of sufficient size and create a seal. Once the object is picked up, it is moved to the pose estimation stage.

Pose estimation is a challenging problem that is required for advanced manipulation. It goes beyond object detection and predicts the object's 6D Pose (3D translation + 3D rotation). For this task we aim to predict the pose of the body of the chicken, no matter where the wings or legs are positioned. Traditional approaches for Pose Estimation operate on raw pointclouds or compute handcrafted local features and perform correspondence grouping. These methods are very sensitive to object variability, lighting conditions, sensor noise, and hyperparameter settings. More recent Deep Learning-based methods operate on RGB or RGB-D inputs and learn the pose estimation task as a regression or classification task \citep{xiang_posecnn:_2017}.

We evaluate two pose estimation approaches: the augmented autoencoder from \citep{sundermeyer_implicit_nodate} and a direct regression approach. The direct regression method combines the ideas from some recent data-driven approaches. Namely, we parameterize the orientation as a quaternion (with respect to the camera frame) and perform regression based on the RGB input. To make the model more robust, we generate a synthetic dataset of different poses rendered using a CAD model and random backgrounds, with extreme augmentations applied on the resulting images, following the procedure in \citep{sundermeyer_implicit_nodate}. Finally, the estimated pose is transformed into the robot frame based on the extrinsic calibration, and the object, treated as the end effector, is sent to the canonical pose. The main contributions of this paper are: 6D Pose Estimation that is robust to the deformable nature of poultry products, and a gripper design and grasping approach capable of handling deformable objects. 
\section{Related work}

\subsection{Robotic Bin Picking}
Robotic bin picking is a challenging yet important technique that could potentially automate many tasks in industrial assembly and processing. What makes robotic bin picking so difficult is the requirement of many sub-tasks to work robustly in concert. The typical bin picking pipeline consists of perception, grasp prediction, and robot control \citep{Shao2019SuctionGR}. The perception system should, ideally, be able to identify an object's position and pose with respect to the robot. In environments with many parts scattered randomly about the scene, the perception system is expected to still function amidst clutter and occlusions from other objects. Traditional grasp prediction entails analysis of force and form closure when using multi-fingered or parallel-finger gripper designs \citep{LiuNFinger}. This requires modelling of the end effector contact friction and object, making it inappropriate for objects that are non-rigid. Data-driven approaches have been explored to mitigate the need for this complex analysis. One particular work \citep{Redmon2015RealtimeGD} explores an end-to-end approach to grasp prediction that produces a grasp for parallel plate robotic grippers. Another contribution  \citep{Shao2019SuctionGR} presents a self-supervised approach to robotic bin picking that allows the robot to learn proper picking points in a cluttered bin; however, the authors only report success in a simulated environment. Approaches like these eliminate the need for grasp pose estimation and modelling since the network learns the mapping from input image to grasp. Given a goal pose, the robot controller should drive the joints of the manipulator such that the end effector reaches the goal pose. 

\subsection{Pose Estimation}
Traditional approaches to pose estimation commonly rely on matching handcrafted features based on pointcloud or RGB-D data and require a 3D model for matching. The basic pointcloud-based approaches, often used for pointcloud registration (which corresponds to computing relative 6D pose between two pointclouds), work by model alignment algorithms, such as ICP \citep{besl1992method} for small misalignments or a global search alternative 4-PCS \citep{aiger20084}. Those algorithms are very sensitive to the hyperparameter settings, pointlcouds content, resolution, and noise. Other methods compute local feature descriptors on the RGB-D input and attempt to match it at the test time. LINEMOD \citep{hinterstoisser2012model} is a popular approach that works by template matching. It computes visual and geometric features (such as gradients and normals) for various viewpoints and then finds the closest match to the view on the test image. Those traditional approaches have high sensitivity to hyperparameter choices, lighting conditions, occlusions, etc. More recent Deep Learning approaches aim to learn the relevant features directly from data. While there has been progress in learning directly on 3D data \citep{qi2017pointnet}, most methods rely on more established Convolutional Neural Networks for image-like inputs. Recent methods, such as PoseCNN \citep{xiang_posecnn:_2017}, have a feature extractor followed by regression networks for the orientation and translation components respectively. While formulating the problem as a regression or classification (by discretizing the orientation space) is most common, other approaches were implemented. The approach presented by \citep{sundermeyer_implicit_nodate} trains an augmented autoencoder network that takes a noisy input and aims to reconstruct a clean image of the object while preserving its orientation. At test time the orientation is predicted by matching the embedding produced by the encoder to a dictionary of embeddings covering all the views of the object. It should be noted that the majority of data-driven approaches only produce a course pose estimate and are typically followed by some form of ICP to produce a refined estimate.

\section{Methodology}

We leverage the power of deep neural networks to handle both our pose estimation and bin picking tasks. We experiment with two techniques for pose estimation. One such approach is that of supervised learning via direct rotation regression. We demonstrate an automated data collection tool that provides us with datasets for regression. The other approach is based on a semi-supervised approach \citep{sundermeyer_implicit_nodate}. We are able to avoid the need for complex gripper design and grasp analysis with our approach of direct grasp point regression.

\subsection{Pose Estimation}

\begin{figure}
\begin{center}
\includegraphics[width=8.4cm]{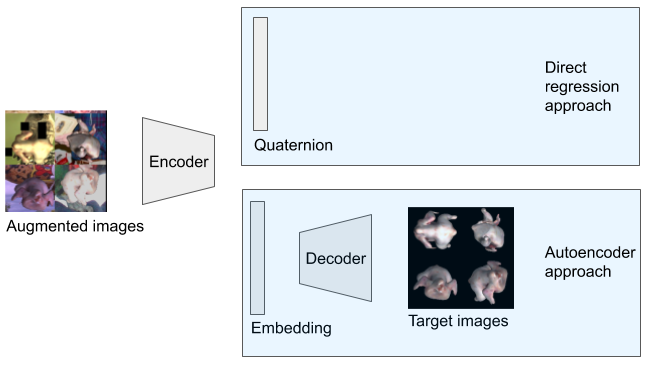}    
\caption{The architectures used for pose estimation. Both direct regression and autoencoder architectures were evaluated.} 
\label{fig:pose_arch}
\end{center}
\end{figure}

\begin{figure}
\begin{center}
\includegraphics[width=8.4cm]{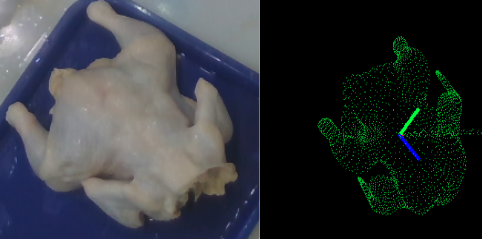}    
\caption{An example output of the pose estimation. The image on the right depicts the local reference frame we have assigned to the bird.} 
\label{fig:pose_example}
\end{center}
\end{figure}

The goal of the Pose Estimation problem is to predict a transformation T = [Rt] that describes the pose of the object with respect to a reference frame. In our case, we assign a local reference frame to the object and then predict the pose of that frame in the camera's optical frame. The local reference frame can be viewed in Fig. \ref{fig:pose_example}. The rotation component of the pose can be expressed using several representations, namely, Euler angles, Rotation matrices, and quaternions. The Euler angle convention gives an intuitive representation of three local axis rotations: yaw, pitch, roll. It suffers, however, from the fact that if the rotations are not bounded within particular regions, there exists infinitely many angle rotations that describe the same transform. Rotation matrices parameterize SO(3) rotations as a 3 x 3 rotation matrix. This convention is useful for visualization, as it directly rotates every point on a model to the new pose. Unfortunately, using nine values to characterize SO(3) rotations requires six implicit constraints in order to describe independent rotations. Due to this, not all members of the set of 3 x 3 rotation matrices are independent. The quaternion convention expresses rotation as a vector and scalar. The only constraint on the representation is that the quaternion has a norm of one; in other words, the quaternion must be a unit vector in order to express a valid rotation. The fact that the representation only imposes a single constraint makes it particularly attractive for non-linear optimization. Since we define the task as a regression problem, we use the quaternion representation, thus the goal of the network is to produce a unit vector of length four. The pose estimation architecture is depicted in Fig. \ref{fig:pose_arch}.

The loss function for the problem should capture the distance between two rotations represented by quaternions; hence, the most appropriate metric is geodesic distance. Let $q_{pred}$ be the predicted quaternion and $q_{gt}$ the ground truth quaternion, then $q_{rel} = \langle q_{pred}, q_{gt} \rangle$ represents the relative rotation between the two orientations. Recall that the scalar component of the quaternion is defined as $w=cos(\theta/2)$. Thus, we can obtain the equation for the geodesic distance between two rotations expressed as quaternions: 

\begin{equation} \label{eq:geodesic}
\theta = 2 arccos(|\langle q_{pred}, q_{gt} \rangle|).
\end{equation}

This is the metric we use to evaluate the accuracy of the pose estimation; however, trigonometric functions are very numerically unstable during the optimization. As such, during the training we use another loss function that acts as a distance metric. Below is the distance metric used as a loss function:
\begin{equation} \label{eq:loss}
dist = \log (1 - |\langle q_{pred}, q_{gt} \rangle| + \epsilon).
\end{equation}

Generating the dataset of real images with manual pose labels is a very labor-intensive process, thus we utilize two automated data sources. First, we implemented the automatic data collection tool. In this case, the object is rigidly attached to a robot arm that was pre-calibrated to the cameras. The tool uses the robot's joint angles to compute the object's pose at the time the images are captured and saves it as pose labels in the camera frame. The transformation from the robot base to the camera is obtained using Eq. (\ref{eq:tool}). Second, we used the approach to generating the synthetic data from \citep{sundermeyer_implicit_nodate}. In this approach, the object is renderred in different poses using a CAD model. During the training, significant augmentations are applied on the input: translation and scale, color jitter, pixel occlusions, and random background. Those augmentations have significant benefits: it allows the model to generalize from the synthetic to the real domain and make the model robust to the noise in the sensor and object detection component of the pipeline. Furthermore, we believe that the augmentations push the network to focus on the overall shape rather than the texture of the object in different poses. A recent work by \cite{geirhos_imagenet-trained_2018} demonstrated that neural networks are biased toward the texture when making predictions and that introducing significant color and texture augmentations help nudging the network to place  emphasis on the more global shape structure. High variability of natural objects makes texture a poor cue for the pose estimation, thus increased emphasis on shape likely helps to make the model more robust.

\begin{equation} \label{eq:tool}
T^{camera}_{object} = (T^{base}_{camera})^{-1} T^{base}_{wrist} T^{wrist}_{object}.
\end{equation}

Using the metric from Eq. (\ref{eq:geodesic}) we evaluated the performance of the regression on the test set (of new birds and poses) collected using the automatic data collection tool. This resulted in a mean geodesic error of 6.582 degrees, however we observed that this performance did not translate well into the experimental setup. It is likely that the images with automatically labelled poses made the model biased to those conditions and made the pose estimation in the test conditions more difficult. Thus, for the experiments with robot the pose estimation was performed using the augmented autoencoder architecture \citep{sundermeyer_implicit_nodate} that only uses the synthetic data. In this approach, the model does not regress to the rotation directly, but instead represents it implicitly as an embedding produced by the encoder. During the training this embedding is fed into the decoder to reconstruct the original image without the added augmentations and computes the pixel-wise MSE loss. During the inference the embedding is compared to a dictionary of embeddings associated with the object's poses on a unit sphere, the pose corresponding to the closest embedding is then predicted by the model. For more details on the autoencoder approach refer to the original paper. The comparison of the two architectures is depicted in Fig. \ref{fig:pose_arch}.

One of the challenges of this particular application is that the pose estimator should be robust to the deformations of the object. We believe that the augmented autoencoder approach achieves this mainly due to the extreme random color augmentations used to generate images, as well as the presence of milder geometric augmentations. Additionally, rendering synthetic images enables us to generate a much larger dataset with a variety of poses, which won't be possible if the images were collected and annotated manually. It should be noted that in this application we are not interested in modelling the deformations of individual parts, however adding deformations based on articulated joints to the image rendering stage is likely to result in an even more robust model.

\subsection{Bin Picking} \label{sec:picking}

The objective of the Bin Picking task is to pick a singulated object from a bin of multiple objects. Our approach to this problem uses a deep learning-based object detector and keypoint predictor in order to predict the picking points and a suction cup gripper used to pick individual birds. Our robotic system consists of a RealSense D435 camera and a Universal Robots UR-5 arm that are calibrated with respect to each other. To perform the calibration, we attach an Aruco tag to the robot as the end effector, where the transformation from the wrist to the tag's origin was previously measured and the transformation from the robot base to the wrist at a given time is available from the driver. Then, the tag's pose in the camera frame is detected and added to the transformation tree. Finally, the transformation from the robot base to the camera is computed using the following transformation chain:

\begin{equation} \label{eq:calib}
T^{base}_{camera} = T^{base}_{wrist} T^{wrist}_{aruco} T^{aruco}_{camera}.
\end{equation}

\begin{figure}[!htb]
\includegraphics[width=8.4cm]{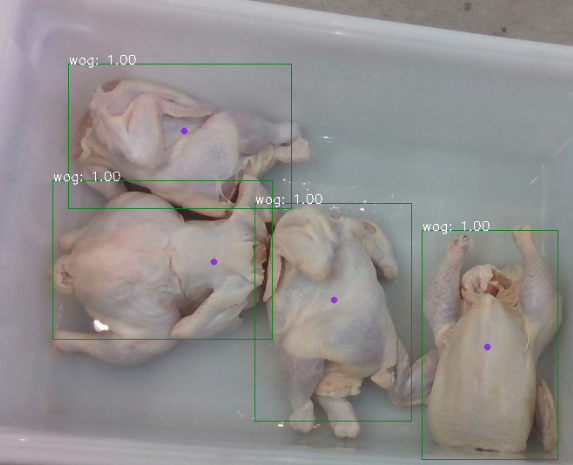}    
\caption{Bin with chickens. Our perception system predicts bounding boxes, picking points, and chicken poses. Bounding boxes and picking points are shown here.} 
\label{fig:example_pred}
\end{figure}

The first step toward a successful grasp is to predict the optimal point of contact. To detect individual objects and predict the picking points for each of them, we use Mask R-CNN architecture \citep{he_mask_2017} with an imagenet-pretrained Resnet101 feature extractor. The training is based on a dataset of 514 images that were manually labelled to produce the ground truth bounding boxes and keypoints. We train this network for 50,000 iterations with an initial learning rate of 0.001. We apply a stepper schedule, decaying the learning rate by 1e-3 at steps 10,000, 30,000, and 40,000. Image color augmentations were employed to artificially supplement the dataset. We train with an 80/20 split and obtain a mAP of 0.874 @ IoU=0.75. At test time, the keypoints are predicted based on the RGB image from the camera and the candidate is chosen based on the confidence score. Then, the pixel coordinate of the point is converted to 3D using the camera intrinsics and the corresponding depth image. Finally, the 3D point is transformed into the robot frame and passed to the controller for manipulation. An example of our detection and keypoint prediction is shown in Fig. \ref{fig:example_pred}.

\subsection{Robot Controls}
\begin{figure}[!htb]
\includegraphics[width=8.4cm]{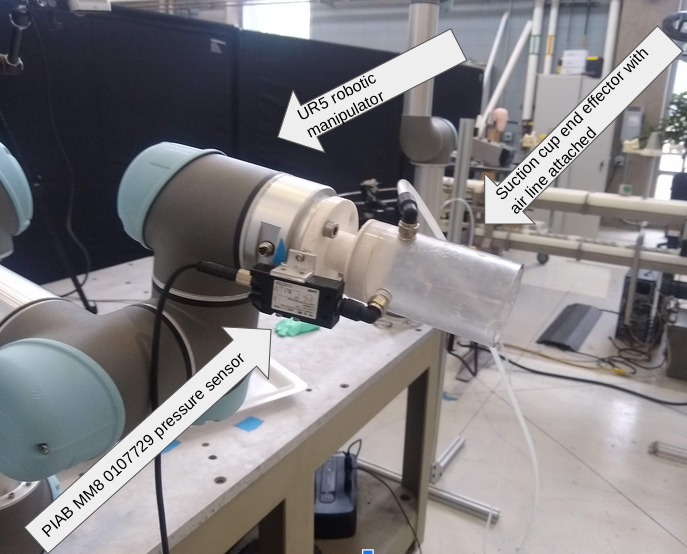}    
\caption{Our robot setup.} 
\label{fig:robot_setup}
\end{figure}

The UR5 from Universal Robots performs the manipulation in our system. This six degree-of-freedom manipulator has a payload of five kilograms and a reach of 850 millimeters, allowing for a spacious, dexterous workspace. The robot is controlled by a standard velocity controller \citep{slotine1991applied} that operates in Cartesian space. Because the robot is operating in a controlled, obstacle-free environment, we omit any motion planning and simply drive the end effector toward the desired position and orientation. To achieve robust picking, we attach a PIAB MM8 0107729 pressure sensor to the suction cup. When executing a bin pick, the robot first places the suction cup end effector above the keypoint found by the perception system. Once the cup is in this position, the robot lowers the cup until the pressure sensor reports a voltage that exceeds a defined threshold. This voltage is inversely proportional to the pressure that the sensor reads. After sensing this voltage, the manipulator lifts the cup with the chicken attached. Before the chicken is lifted, the perception system predicts a pose in the frame of the camera. Given the pose of the bird in the camera frame and the transformation from the camera to the suction cup, we append the bird to the transformation tree, treating it as a new end effector. After this, the robot can be commanded to send the bird to the canonical pose. Our system can be seen in Fig. \ref{fig:robot_setup}.

\section{Experiments}

\subsection{Bin Picking Task}

To evaluate the performance of the bin picking task, we isolated this step of the pipeline and performed tests. We placed a bin of chickens in the obstacle-free picking area and had the robot attempt to pick up the birds. We evaluated the effectiveness of the robot with the bird in three different poses: breast up, breast down, and breast side. Examples of bird poses are shown in Fig. \ref{fig:example_pred}. To preserve the element of stochasticity and variability found in real poultry processing, we cycle new chickens throughout experiment. It should also be noted that the birds have been kept in chilled ice water prior to their use in experimentation. We observed that bird temperature affects the stiffness of the skin and, consequently, the performance of the suction cup. A pick is considered successful if the bird is lifted from the bin without falling off of the gripper. Table \ref{table:1} shows the success rates of the picking task with respect to each bird pose.

\begin{table}[h!]
\centering
\begin{tabular}{||p{1.3cm} p{1.3cm} p{1.3cm} p{1.3cm} p{1.3cm}||} 
 \hline
 Pose & Breast Up & Breast Down & Breast Side & Total  \\ [0.5ex] 
 \hline\hline
 Attempted & 25 & 25 & 25 & 75\\ 
 Succeeded & 23 & 22 & 19 & 64\\
 Success Rate & 92\% & 88\% & 76\% & 85\%\\[1ex] 
 \hline
\end{tabular}
\newline
\centering
\caption{Bin Picking success rate}
\label{table:1}
\end{table}

\subsection{Canonical Pose Placement Task}

\begin{figure}[!htb]
\includegraphics[width=8.4cm]{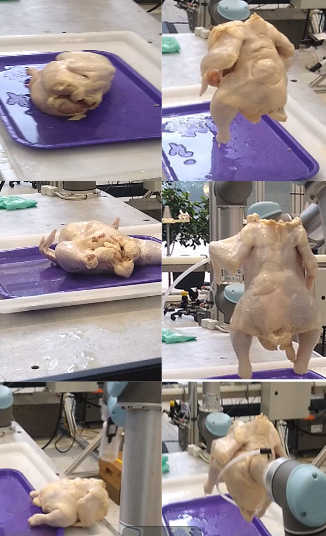}    
\caption{Visual results of placement. The bird starts in an unknown state and is brought to a known configuration. } 
\label{fig:res1}
\end{figure}

The next and final experiment was done to assess the performance of the pose placement task. In this test, we want the robot to pick up a singulated chicken from the picking area, and hold the bird in a consistent pose. Regardless of the starting position and orientation of the bird, the chicken should end up in the same canonical pose. Like the previous experiment, we employ variability in position, orientation, and rotate birds throughout the test. The pose placement is considered successful if the breast of the chicken is facing the picking area and the neck of the chicken is pointing up towards the ceiling. Images of the canonical pose are shown in Fig. \ref{fig:res1}. We evaluate both the augmented autoencoder and the direct regression method in this test. The results for this experiment are given in tables \ref{table:2} and \ref{table:3}.

\begin{table}[h!]
\centering
\begin{tabular}{||p{1.3cm} p{1.3cm} p{1.3cm} p{1.3cm} p{1.3cm}||} 
 \hline
 Pose & Breast Up & Breast Down & Breast Side & Total \\ [0.5ex] 
 \hline\hline
 Attempted & 30 & 30 & 30 & 90\\ 
 Succeeded & 10 & 14 & 5 & 29 \\
 Success Rate & 33\% & 46\% & 16\% & 32\%\\[1ex] 
 \hline
\end{tabular}
\newline
\centering
\caption{Pose Placement Success Rate (Direct Regression)}
\label{table:2}
\end{table}

\begin{table}[h!]
\centering
\begin{tabular}{||p{1.3cm} p{1.3cm} p{1.3cm} p{1.3cm} p{1.3cm}||} 
 \hline
 Pose & Breast Up & Breast Down & Breast Side & Total  \\ [0.5ex] 
 \hline\hline
 Attempted & 30 & 30 & 30 & 90\\ 
 Succeeded & 25 & 28 & 20 & 73 \\
 Success Rate & 83\% & 93\% & 66\% & 81\%\\[1ex] 
 \hline
\end{tabular}
\newline
\centering
\caption{Pose Placement Success Rate (Augmented Autoencoder)}
\label{table:3}
\end{table}

\subsection{Discussion of Experimental Results}
The keypoint prediction gave somewhat consistent results for each of the poses. Occasionally, the keypoint selection would pick a point that was not flat on the chicken, resulting in a failed pick. The pose that yielded the most failed picks was the \textit{Breast Side} pose. Sometimes, the wing of the chicken would be in the way of the suction cup. Even if the keypoint was not directly placed on the wing, the wing would still obstruct the suction cup and prevent a clean seal. We believe these two problems can be eliminated with more training data. As shown in tables \ref{table:2} and \ref{table:3}, the augmented autoencoder vastly outperformed the direct regression for pose placement. The direct regression model exhibited high variance in its predictions. Given slightly shifted images of a chicken in the same pose, the model would predict a highly varying set of quaternions for the same pose. This is likely due to the significant distribution shift between the training and test data. The augmented autoencoder approach demonstrated much better generalization and performed well in predicting the bird pose.

\section{Conclusion}
In this paper, we present an automated system for picking birds and placing them into a canonical pose. The two main contributions of our paper is the picking approach that is capable of detecting the desirable point of contact and adapt to the deforming surface, and the pose estimation approach that generalizes well to the environment and object variability. The experimental evaluation of our robotic system demonstrates that it can achieve high accuracy for both tasks in real-world conditions. However, further refinement of the pose estimation approach is required to make the predictions more consistent for various initial states. Overall, by transforming the poultry products from an unordered state into a known canonical pose, this work suggests the possibility of automating complex tasks in food processing that involve variability and stochasticity.

\begin{ack}
The research presented in this paper was supported by the Agricultural Technology Research Program of the Georgia Tech Research Institute.
\end{ack}

\bibliography{ifacconf}             

\end{document}